\documentclass[10pt, a4paper]{article}

\usepackage{lrec-coling2024} 

\usepackage{listings}

\usepackage{natbib}
\usepackage{multibib}
\makeatletter
\def\@mb@citenamelist{cite,citep,citet,citealp,citealt,citepalias,citetalias}
\makeatother
\newcites{languageresource}{~}

\usepackage{graphicx}
\usepackage{tabularx}
\usepackage{soul}
\usepackage{xcolor}
\usepackage{titlesec}
\titleformat{\section}{\normalfont\large\bfseries\center}{\thesection.}{1em}{}
\titleformat{\subsection}{\normalfont\SmallTitleFont\bfseries\raggedright}{\thesubsection.}{1em}{}
\titleformat{\subsubsection}{\normalfont\normalsize\bfseries\raggedright}{\thesubsubsection.}{1em}{}
\renewcommand\thesection{\arabic{section}}
\renewcommand\thesubsection{\thesection.\arabic{subsection}}
\renewcommand\thesubsubsection{\thesubsection.\arabic{subsubsection}}

\usepackage{xcolor}
\usepackage{hyperref}
 \definecolor{darkblue}{rgb}{0, 0, 0.5}
  \hypersetup{colorlinks=true, citecolor=darkblue, linkcolor=darkblue, urlcolor=darkblue}

\usepackage{xstring}

\usepackage{color}

\usepackage{booktabs}
\usepackage{multirow}
\usepackage{makecell}
\usepackage{amsmath}

\title{On The Adaptation of Unlimiformer for Decoder-Only Transformers}

\name{Kian Ahrabian$^{\bf 1}$$^{\bf \ast}$\thanks{$^{\ast}$ Work done during an internship at Microsoft.}, Alon Benhaim$^{\bf 2}$, Barun Patra$^{\bf 2}$\\{\bf \large Jay Pujara$^{\bf 1}$, Saksham Singhal$^{\bf 2}$, Xia Song$^{\bf 2}$}}

\address{
  $^{1}$ Information Sciences Institute, University of Southern California \\
  $^{2}$ Microsoft \\
  \texttt{ahrabian@usc.edu}, \texttt{\{alonbenhaim,barun.patra\}@microsoft.com} \\
  \texttt{jpujara@isi.edu}, \texttt{\{saksham.singhal,xiaso\}@microsoft.com}
}

\abstract{
One of the prominent issues stifling the current generation of large language models is their limited context length.
Recent proprietary models such as GPT-4 and Claude 2 have introduced longer context lengths, 8k/32k and 100k, respectively; however, despite the efforts in the community, most common models, such as LLama-2, have a context length of 4k or less.
Unlimiformer \cite{bertsch2023unlimiformer} is a recently popular vector-retrieval augmentation method that offloads cross-attention computations to a kNN index.
However, its main limitation is incompatibility with decoder-only transformers out of the box.
In this work, we explore practical considerations of adapting Unlimiformer to decoder-only transformers and introduce a series of modifications to overcome this limitation.
Moreover, we expand the original experimental setup on summarization to include a new task (i.e., free-form Q\&A) and an instruction-tuned model (i.e., a custom 6.7B GPT model).
Our results showcase the effectiveness of these modifications on summarization, performing on par with a model with 2x the context length.
Moreover, we discuss limitations and future directions for free-form Q\&A and instruction-tuned models.
\\ \newline \Keywords{Large Language Models, Decoder-Only Transformers, Retrieval-Augmented Attention}
}

\begin{document}

\maketitleabstract

\section{Introduction}
In recent years, large language models (LLMs) have become critical to many language-based technologies, such as conversational and search systems.
LLMs have shown state-of-the-art performance on sequence-to-sequence downstream tasks such as summarization and question-answering.
However, the performance of these models is bounded by the information that can fit in their context (see Figure \ref{fig:exp:datasets} and Section \ref{exp:eval}).
Despite the efforts in the community~\cite{choromanski2021rethinking,beltagy2020longformer,ivgi-etal-2023-efficient}, most of the common open-source models, e.g., MPT~\cite{MosaicML2023Introducing}, Falcon~\cite{penedo2023refinedweb}, and LLama-2~\cite{touvron2023llama}, have a context length of 4096 or less.
As such, efficiently overcoming this limitation would allow a broader and fairer adaptation of LLMs while increasing their performance across benchmarks.

\begin{figure}[t]
\centering
\includegraphics[width=\columnwidth]{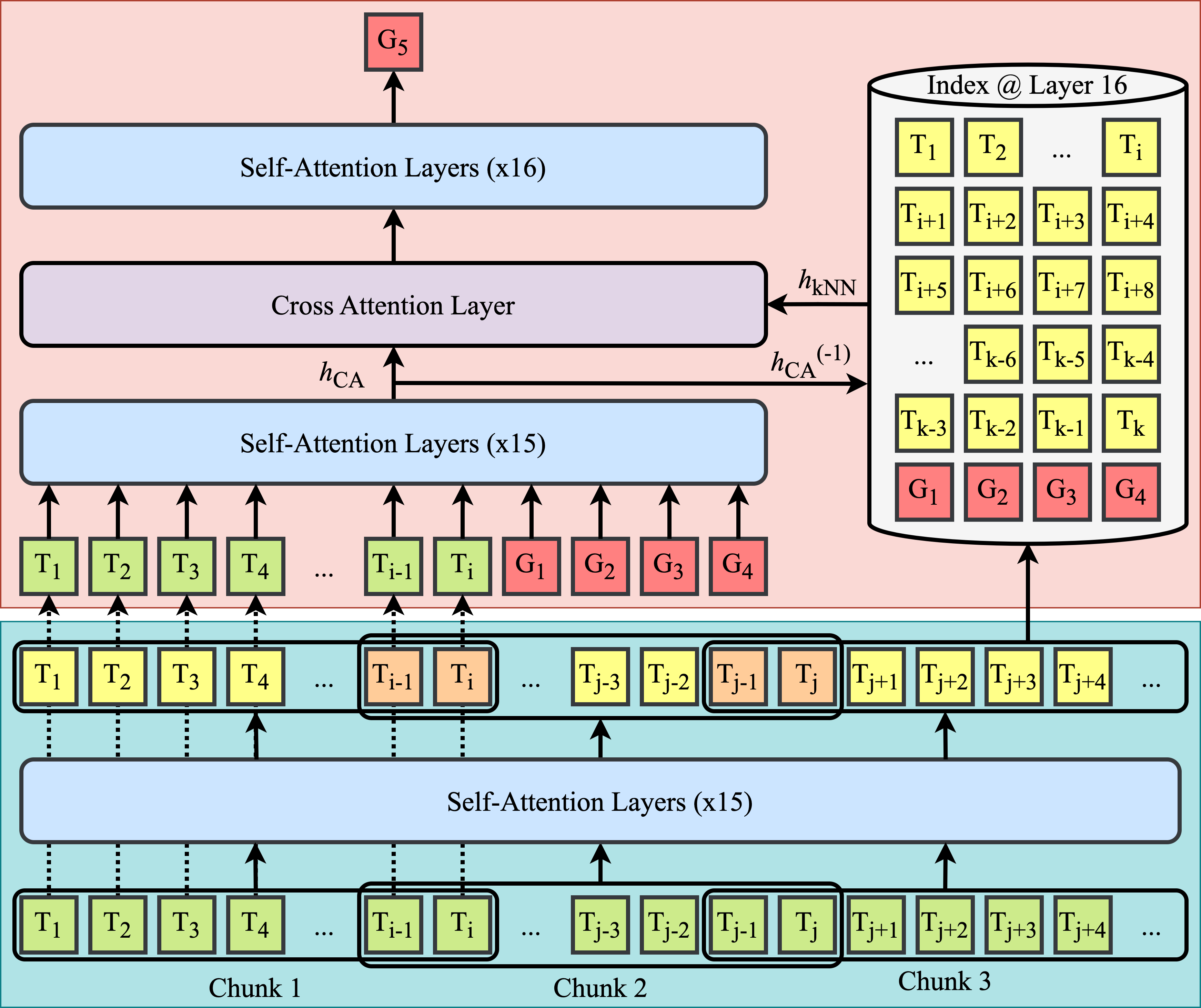}
\caption{
Overview of an example of the adapted decoder-only model where only a single layer (e.g., 16th) uses cross-attention.
Here, $T_i$ and $G_i$ represent the original context and generated tokens. 
The cyan section encapsulates the first pass to create the indices, while the pink section illustrates the second pass to generate sequences.
Note that the first chunk appears in both the input and the index by design, keeping the input the same in all variations of our experiments.}
\label{fig:intro:over}
\end{figure}

In general, most of the existing methods for extending the contextual information in LLMs focus on one of the following approaches:
1) extending positional embeddings through extrapolation or interpolation~\cite{press2022train,sun-etal-2023-length}, 2) introducing recurrence in the transformer~\cite{hutchins2022blockrecurrent,yang2019xlnet}, 3) introducing sparsity in the attention mechanism~\cite{beltagy2020longformer,zaheer2020big}, and 4) augmenting the transformer with a vector-retrieval module~\cite{rubin2023long}.
One popular approach that has gained much traction recently for extending contextual information to unlimited inputs (theoretically) is Unlimiformer \cite{bertsch2023unlimiformer}.
Unlimiformer is a vector-retrieval augmentation method that offloads the cross-attention computations to a kNN index and can wrap \textit{any existing encoder-decoder model}.
However, its incompatibility with decoder-only models is a significant shortcoming.

\noindent\paragraph{Contributions:}
In this work, we present a set of modifications to overcome this limitation of the Unlimiformer architecture, adapting it to the decoder-only models (see Figure \ref{fig:intro:over}).
These modifications consist of 1) modifying the cross-attention formulation to include information fusion, 2) updating the index creation procedure, 3) addressing the index staleness problem, and 4) adapting the chunk encoding procedure to causal attention (see Section \ref{sec:meth}).
Moreover, we introduce a new evaluation setting and present our experiments on four long-document datasets across two tasks: summarization and free-form Q\&A.
Our experiments show that our modifications improve summarization datasets, performing on par with a model with 2x context length.
We also discuss the limitations and future directions for free-form Q\&A and instruction-tuned models.

\section{Related Works}
Many prior works, such as Linformer~\cite{wang2020linformer} and Reformer~\cite{Kitaev2020Reformer}, have been focused on creating more efficient transformers.
\citet{10.1145/3530811} present a comprehensive study on these models.
Moreover, there have been efforts to accelerate dense attention calculations by crafting IO-aware CUDA kernels~\cite{dao2022flashattention, dao2023flashattention}.
Recently, \citet{rubin2023long} have introduced another retrieval-augmented attention model for decoder-only transformers; however, unlike Unlimiformer, this model does not work in zero-shot settings.
Finally, there have been attempts to break away from attention-based models entirely through linear RNNs~\cite{peng2023rwkv} or convolutions~\cite{poli2023hyena}.

\section{Methodology}
\label{sec:meth}
In summary, Unlimiformer consists of three main steps:
1) split the input into overlapping digestible chunks,
2) encode each chunk and store the hidden states of the middle-half tokens in a kNN index,
and 3) approximate the dense attention in the decoder using a subset of hidden states retrieved from the index.
In this section, we discuss our changes to make Unlimiformer compatible with decoder-only models\footnote{Concurrently, \citet{bertsch2023unlimiformer} have released an implementation for decoder-only models. Appendix \ref{app:unlim} details the differences in our adaptation.}.

\subsection{Cross-Attention}
\label{sec:meth:ca}
Formally, in Unlimiformer, the dot-product part of the cross-attention mechanism in decoder layers is approximated as
\begin{equation}
    QK^T \approx (h_dW_qW_k^T)h_e^T
\end{equation}
where $h_d$ is the decoder's hidden states, and $h_e$ is the retrieved hidden states that maximize $(h_d^{(-1)}W_qW_k^T)h_e^T$.

In decoder-only transformers, due to the absence of a natural encoding/decoder splitting layer, we can arbitrarily choose any layer to use cross-attention instead of self-attention.
This allows us to use various simple or complex patterns for the set of cross-attention layers, $\mathcal{L}$ (e.g., $\mathcal{L} = \{16\}$ in Figure \ref{fig:intro:over}).
Appendix \ref{app:hyper} details how these patterns are tuned as hyperparameters of the model.

Let $h_{\text{CA}}$ be the input to a cross-attention layer.
Based on $h_{\text{CA}}^{(-1)}$, we can retrieve the relevant vectors, $h_{\text{kNN}}$, from the index.
Similar to the memory transformer~\cite{burtsev2020memory}, by fusing $h_{\text{kNN}}$ and $h_{\text{CA}}$ we form the new query matrix $h_{q}$, and the new key-value matrix, $h_{kv}$ as
\begin{align}
    h_{q} &= [h_{\text{kNN}}[\alpha_{q}:]; h_{\text{\text{CA}}}[\beta_{q}:]] \\
    h_{kv} &= [h_{\text{kNN}}[\alpha_{kv}:]; h_{\text{\text{CA}}}[\beta_{kv}:]]
\end{align}
where $\alpha_{q}$ and $\beta_{q}$ ($\alpha_{kv}$ and $\beta_{kb}$) control the retention sizes of the retrieved and input vectors in the query (key-value) matrix.
This fusion scheme gives us more flexibility on what information is processed in the attention mechanism.
Given $h_{q}$ and $h_{kv}$, we can approximate the dot-product part of the cross-attention as
\begin{equation}
    QK^T \approx (h_{q}W_q)(h_{kv}W_k)^T \  .
    \label{eq:ca}
\end{equation}
Similar to the original approach, $W_q$ and $W_k$ are head-specific but use the same index.


\subsection{kNN Indices}
Without separate encoder/decoder modules, operating with only one index for the whole model is impossible.
This is because each layer in a decoder-only model attends to the output of its previous layer, in contrast to the decoder layers in encoder-decoder models that attend to the encoder's output.
Consequently, if we arbitrarily use a specific layer's output for building our index, we create a distributional mismatch between the expected input and actual inputs of future layers.
Hence, we must create a separate index for each cross-attention layer to overcome this issue.
However, this approach requires much more memory and computation cost to build the indices.
For example, in our models, it could add up to $|\mathcal{L}|$ times the cost of having a single index.
Section \ref{sex:limit} discusses the potential ways of mitigating this issue.
In our experiments, we only tune our models to use at most three cross-attention layers, i.e., $|\mathcal{L}| = 3$.

\subsection{Index Staleness}
One of the potential issues of a static index is staleness.
Specifically, since Unlimiformer approximates dense attention, without updating the index, we would lose the information from the newly generated tokens, which might lead to incoherent outputs.
For example, assume we have a model with a context length of $N$.
In this scenario, at generation step $N + 2$, the input to the model would be the last $N$ generated tokens, effectively discarding the first generated token as it is also absent from the index.
To fix this problem, at each generation step and each layer, we add $h_{\text{CA}}^{(-1)}$ to its respective index.
This ensures the addition of the most recent token and keeps the indices from going stale.
Appendix \ref{app:abl:stale} presents an ablation study that showcases the effectiveness of this simple change.

\subsection{Chunks Encoding}
In contrast to encoder-decoder models, decoder-only transformers use causal (unidirectional) attention.
This difference means that a token has seen enough contextual information if a certain number of tokens are behind it.
As a result, instead of only storing the hidden states of the middle half tokens, we can keep all the non-overlapping ones.
This will allow us to be slightly more efficient when processing long documents.
Note that only the first instance of overlapping tokens is added to the index, as illustrated by orange tokens in Figure \ref{fig:intro:over}.

\section{Experimental Setup}

\begin{figure}[t]
\centering
\includegraphics[width=0.85\columnwidth]{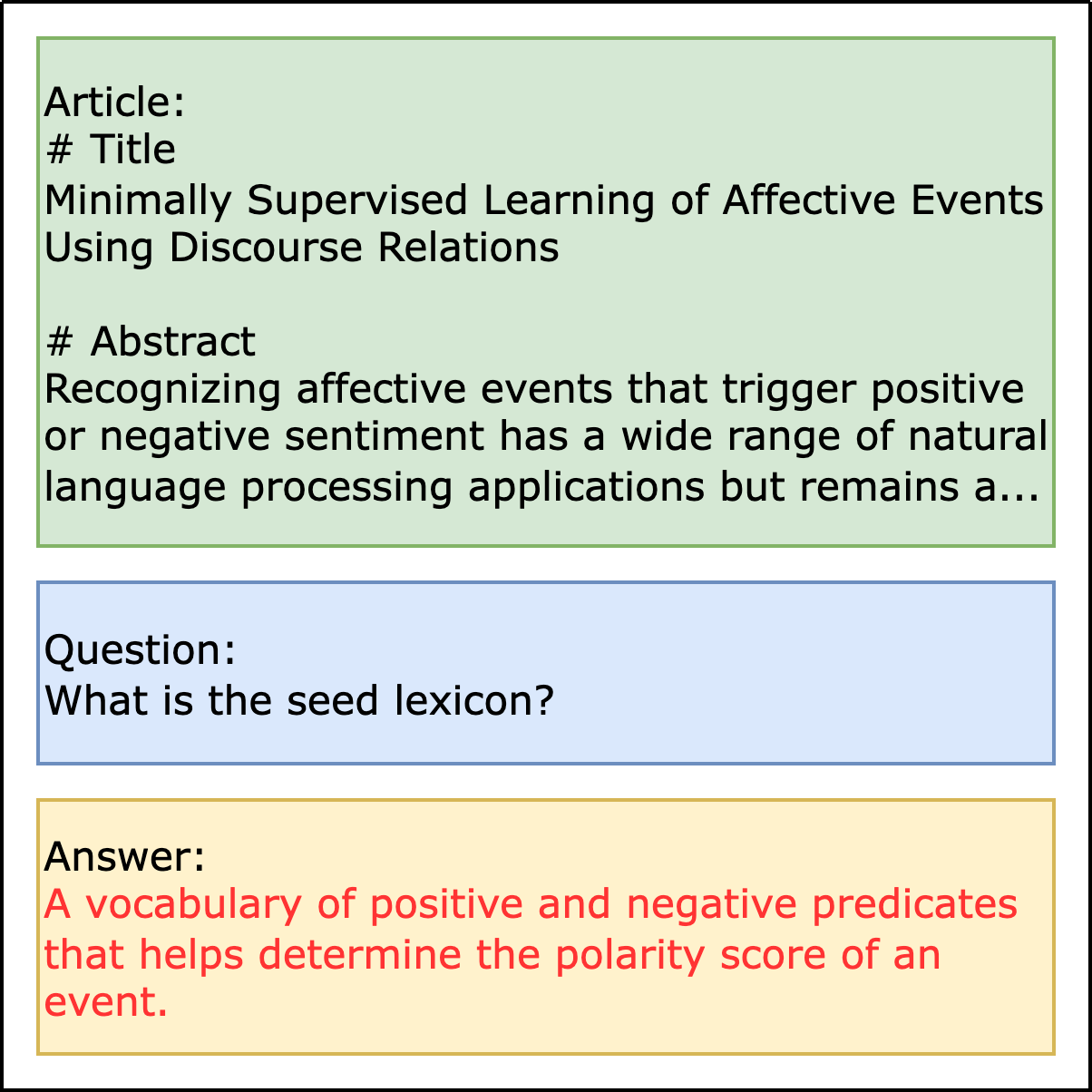} 
\caption{A sample free-form Q\&A prompt. The article section consists of the truncated version of the full article that fits in the context. The question section is always fully included.}
\label{fig:exp:prompt}
\end{figure}

\begin{table}[t]
\centering
\resizebox{\columnwidth}{!}{
    \begin{tabular}{cccc}
        \toprule
        \textbf{Dataset} & \textbf{Metrics} & \textbf{\#Samples} & \textbf{Avg \#Tokens} \\ \midrule
        GovReport (GAO) & \multirow{2}{*}{\makecell{ROUGE-\{1,2,L\},\\METEOR}} & 611 & 11395 \\
        BookSum & & 46 & 164695 \\ \midrule 
        NarrativeQA & \multirow{2}{*}{Token F1} & 10557 & 76433 \\
        Qasper & & 1372 & 4836 \\ \bottomrule
    \end{tabular}
}
\caption{Statistics of the datasets. The average number of tokens is obtained using the GPT-4 tokenizer: \url{https://github.com/openai/tiktoken}.}
\label{tab:exp:datasets}
\end{table}

\subsection{Datasets \& Tasks}
For our experiments, we use datasets from the two tasks of summarization and free-form Q\&A, chosen due to the existence of long-document benchmarks~\cite{shaham2023zeroscrolls}.
Specifically, for summarization, we use GovReport (GAO)~\cite{huang-etal-2021-efficient} and BookSum~\cite{kryscinski-etal-2022-booksum}, and for free-form Q\&A, we use NarrativeQA~\cite{kocisky-etal-2018-narrativeqa} and Qasper~\cite{dasigi-etal-2021-dataset}.
Moreover, we tune the hyperparameters on the validation sets and report the results on the test sets.
All the experiments are carried out in a zero-shot setting.
Table \ref{tab:exp:datasets} presents the statistics of these datasets.

\begin{figure}[t]
\centering
\includegraphics[width=0.9\columnwidth]{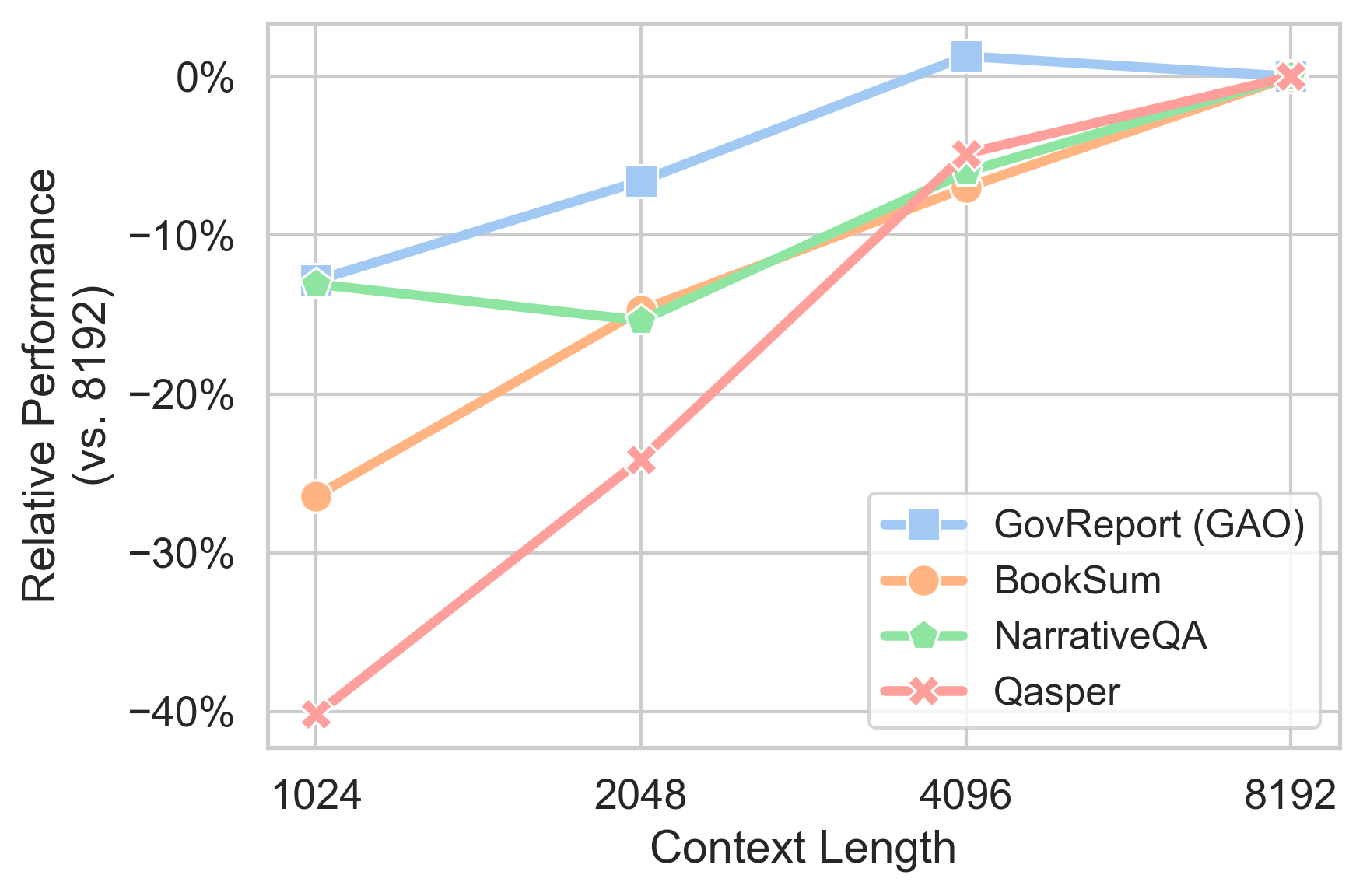} 
\caption{Performance comparison of GPT-Summ on GovReport (GAO) and BookSum, and GPT-Inst on NarrativeQA and Qasper at varying context lengths.}
\label{fig:exp:datasets}
\end{figure}

\subsection{Models}
We used two distinct base models to evaluate our modifications: 1) \textbf{GPT-Summ:} a finetuned summarization model and 2) \textbf{GPT-Inst:} an instruction-tuned model.
To better understand the impact of our approach compared to dense attention, in contrast to prior works, we pre-train both variants of the models with a sequence length of 8192 using the same architecture as the GPT-3 6.7B model~\cite{brown2020language}, with the addition of RoPE embeddings~\cite{su2021roformer}.

\begin{figure*}[t]
\centering
\includegraphics[width=\textwidth]{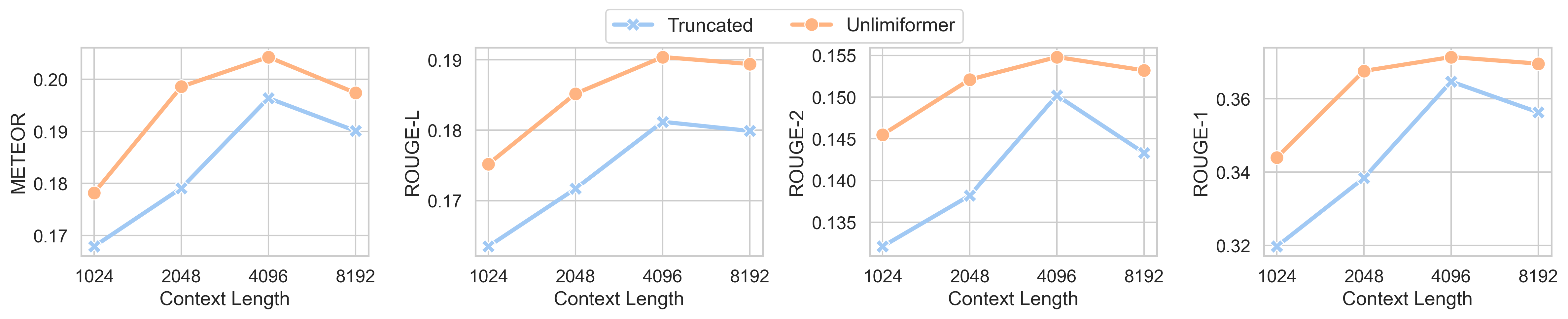} 
\caption{METEOR and ROUGE-\{1,2,L\} achieved by GPT-Summ on the GovReport (GAO) dataset.}
\label{fig:res:govr}
\end{figure*}

\begin{figure*}[t]
\centering
\includegraphics[width=\textwidth]{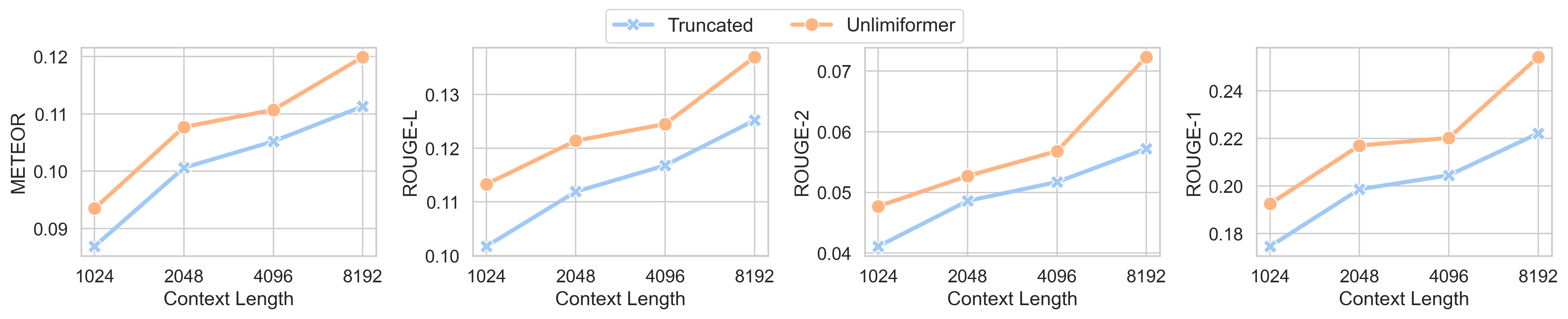} 
\caption{METEOR and ROUGE-\{1,2,L\} achieved by GPT-Summ on the BookSum dataset.}
\label{fig:res:book}
\end{figure*}

\subsection{Evaluation Setup}
\label{exp:eval}
The original Unlimiformer paper presents two main experimental comparisons: 1) $\text{BART}_\text{base}$ vs. $\text{BART}_\text{base} + \text{Unlimiformer}$ and 2) $\text{BART}_\text{base} + \text{Unlimiformer}$ vs. SLED~\cite{ivgi-etal-2023-efficient} and Longformer~\cite{beltagy2020longformer}.
These comparisons showcase the effectiveness of the proposed model; however, one missing crucial evaluation setup is the comparison to the same base model with longer context lengths, e.g., $\text{GPT-Summ}[2048]$ vs. $\text{GPT-Summ}[1024] + \text{Unlimiformer}$.
Such a setup provides more insight into how efficiently Unlimiformer uses the extra information provided through the kNN index.
In this work, we focus on this setting by restricting the context length of the base model to 1024, 2048, 4096, and 8192 tokens and then comparing them to variations equipped with Unlimiformer.
To ensure that our models and datasets showcase meaningful differences across context lengths in such a setup, we ran experiments on the base models using the validation sets, estimating the performance changes as contextual information increased.
Figure \ref{fig:exp:datasets} illustrates the results of our experiments.
Moreover, Appendix \ref{app:abl:info} presents a case study to ensure the performance disparity in Figure \ref{fig:exp:datasets} is an artifact of the datasets, not a deficiency in the models.

\subsection{Prompt Structure}
For the free-form Q\&A datasets and the instruction-finetuned model, we opted for a simple three-part (Article, Question, and Answer) template.
Figure \ref{fig:exp:prompt} illustrates an example of the prompt structure.
We also investigated ZeroScrolls's prompt template~\cite{shaham2023zeroscrolls}, but since we did not notice any significant difference in the performance, we continued with the more straightforward template.

\begin{figure}[t]
\centering
\includegraphics[width=\columnwidth]{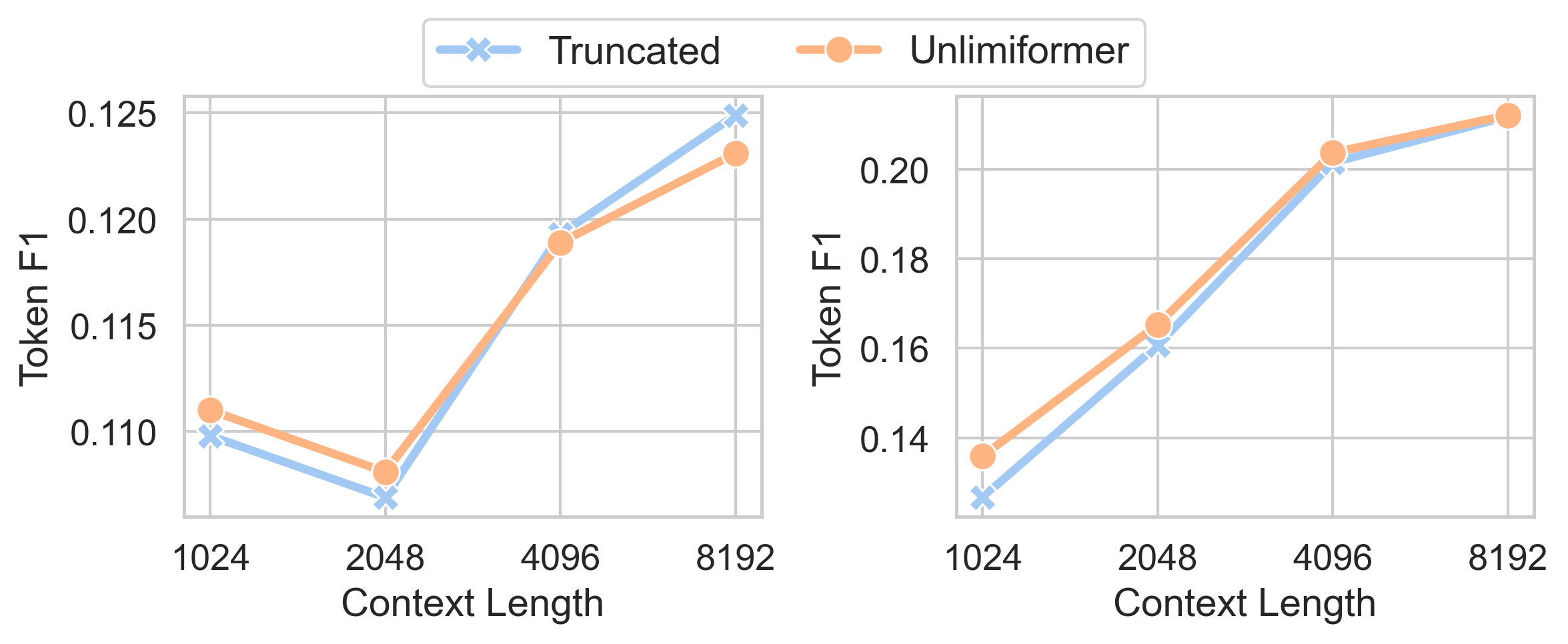}
\caption{Token F1 achieved by GPT-Inst on NarrativeQA (left) and Qasper (right) datasets.}
\label{fig:res:nqa_qas}
\end{figure}

\subsection{Metrics}
For summarization, we report ROUGE-\{1,2,L\}~\cite{lin-2004-rouge} and METEOR~\cite{banerjee-lavie-2005-meteor} as 1) they are standard metrics and 2) have shown good correlation to expert annotations~\cite{tacla00373}.
Moreover, in our early experiments, we investigated BERTScore~\cite{bert-score}; however, similar to \citet{bertsch2023unlimiformer}'s findings, we observed a lack of distinguishing power among the lengthy summaries, even when other metrics and manual inspection showed improvements.
For free-form Q\&A, we used the standard token F1 metric, similar to the ZeroScrolls benchmark.

\section{Results}
\subsection{Summarization}
Figures \ref{fig:res:govr} and \ref{fig:res:book} present our experimental results using the GPT-Summ model on the GovReport (GAO) and BookSum datasets, respectively.
As evident from these results, adding our modifications improves the model's performance to a 2x level (e.g., $ \text{GPT-Summ}[1024] + \text{Unlimiformer} \approx \text{GPT-Summ}[2048]$) on the GovReport (GAO) dataset.
Similarly, we observe significant improvements in the BookSum dataset.
These results showcase the effectiveness of our proposed modifications.

\subsection{Free-Form Q\&A}
\label{sec:res:qa}
Figure \ref{fig:res:nqa_qas} illustrates our experimental results using the GPT-Inst model on the NarrativeQA and Qasper datasets.
Although we can observe some improvements in both Qasper and (mostly) NarrativeQA, they are less significant than the summarization datasets' results.
Given the prompt-based approach used for the free-from Q\&A task (see Figure \ref{fig:exp:prompt}), the insignificant improvements could be an artifact of the instruction-tuned model being too biased toward the input, making it insensitive to the added information in the cross-attention layers.
These results present exciting opportunities to investigate such models in future studies.

\section{Ablations}

\begin{table}[t]
\centering
\small
    \begin{tabular}{cccc}
        \toprule
        \textbf{Variation} & \textbf{Rouge-1} & \textbf{Rouge-2} & \textbf{Rouge-L} \\ \midrule
        w/o & 0.4247 & 0.1734 & 0.2512 \\
        w/ & \textbf{0.4263} & \textbf{0.1744} & \textbf{0.2523} \\ \bottomrule
    \end{tabular}
\caption{Effect of adding newly generated tokens to the index on the performance of the model.}
\label{tab:abl:stale}
\end{table}

\begin{table}[t]
\centering
\resizebox{\columnwidth}{!}{
    \begin{tabular}{ccccc}
        \toprule
        \textbf{Chunk} & \textbf{\makecell{Min Evidence\\> 1024 (\%)}} & \textbf{\makecell{Min Evidence\\> 2048 (\%)}} & \textbf{\makecell{Min Evidence\\> 4096 (\%)}} & \textbf{\makecell{Min Evidence\\> 8192 (\%)}} \\ \midrule
        Train & 66.37 & 47.79 & 19.26 & 1.42 \\
        Valid & 67.77 & 47.03 & 15.30 & 0.87 \\
        Test & 68.22 & 46.02 & 13.71 & 0.75 \\ \bottomrule
    \end{tabular}
}
\caption{Percentage of answers with minimum evidence position outside a given context length in Qasper.}
\label{tab:abl:info}
\end{table}


\subsection{Index Staleness}
\label{app:abl:stale}
To study the effect of index staleness, we experiment with an internal summarization dataset consisting of samples of up to 7k tokens using a model with a context length of 2048 and a generation limit of 700 tokens.
Table \ref{tab:abl:stale} presents the result of our experiments.
Although the generation length is still way under the 2048 limit, we can see a slight positive improvement in the performance, showcasing the usefulness of this simple addition with almost no cost.
Moreover, we believe the performance boost will increase as the generation length increases.

\subsection{Contextual Information}
\label{app:abl:info}
In the Qasper dataset, we have access to a set of evidence for each answer.
Hence, we can calculate the percentage of answers that all of their evidence falls out of the range of a specific context length.
Table \ref{tab:abl:info} presents these numbers for different context lengths.
These numbers are consistent with the improvements in Figure \ref{fig:exp:datasets}, which showcase the validity of our models.

\section{Conclusion}
This work presented a set of changes to adapt the Unlimiformer architecture to decoder-only models.
We evaluated these changes with a new set of experiments, showcasing their effectiveness, especially on summarization datasets.
We also identified a failure case that warrants further investigations in future studies.
We hope this work paves the way for the broader use of this architecture.

\section{Limitations}
\label{sex:limit}
\paragraph{Query Bias}
Since both Unlimiformer and our approach use a specific query vector to retrieve hidden states from the index, the retrieval process becomes biased on the query vector.
In Unlimiformer, this vector is $h_{d}^{(-1)}$, which is calculated by attending to the generated tokens.
In our approach, this vector is $h_{\text{CA}}^{(-1)}$, which is calculated by attending to the whole input, i.e., original context + generated tokens.
This dependence on the original context potentially reduces the expected performance gains when external indices are used.
Moreover, it could partially explain the lack of significant improvements in Section \ref{sec:res:qa}.

\paragraph{Latency}
The main setback of having many indices is the increase in latency.
To alleviate this problem, we could 1) use approximate indices and/or 2) use indices that operate on GPU to remove the CPU-GPU transfer overhead.

\section{Ethical Considerations}
This paper does not have any ethical considerations


\nocite{*}
\section{Bibliographical References}
\bibliographystyle{lrec_natbib}
\bibliography{lrec-coling2024-example}

\begin{thebibliography}{36}
\expandafter\ifx\csname natexlab\endcsname\relax\def\natexlab#1{#1}\fi

\bibitem[{Banerjee and Lavie(2005)}]{banerjee-lavie-2005-meteor}
Satanjeev Banerjee and Alon Lavie. 2005.
\newblock \href {https://aclanthology.org/W05-0909} {{METEOR}: An automatic
  metric for {MT} evaluation with improved correlation with human judgments}.
\newblock In \emph{Proceedings of the {ACL} Workshop on Intrinsic and Extrinsic
  Evaluation Measures for Machine Translation and/or Summarization}, pages
  65--72, Ann Arbor, Michigan. Association for Computational Linguistics.

\bibitem[{Beltagy et~al.(2020)Beltagy, Peters, and
  Cohan}]{beltagy2020longformer}
Iz~Beltagy, Matthew~E Peters, and Arman Cohan. 2020.
\newblock Longformer: The long-document transformer.
\newblock \emph{arXiv preprint arXiv:2004.05150}.

\bibitem[{Bertsch et~al.(2023)Bertsch, Alon, Neubig, and
  Gormley}]{bertsch2023unlimiformer}
Amanda Bertsch, Uri Alon, Graham Neubig, and Matthew~R Gormley. 2023.
\newblock Unlimiformer: Long-range transformers with unlimited length input.
\newblock \emph{arXiv preprint arXiv:2305.01625}.

\bibitem[{Brown et~al.(2020{\natexlab{a}})Brown, Mann, Ryder, Subbiah, Kaplan,
  Dhariwal, Neelakantan, Shyam, Sastry, Askell, Agarwal, Herbert-Voss, Krueger,
  Henighan, Child, Ramesh, Ziegler, Wu, Winter, Hesse, Chen, Sigler, Litwin,
  Gray, Chess, Clark, Berner, McCandlish, Radford, Sutskever, and
  Amodei}]{NEURIPS2020_1457c0d6}
Tom Brown, Benjamin Mann, Nick Ryder, Melanie Subbiah, Jared~D Kaplan, Prafulla
  Dhariwal, Arvind Neelakantan, Pranav Shyam, Girish Sastry, Amanda Askell,
  Sandhini Agarwal, Ariel Herbert-Voss, Gretchen Krueger, Tom Henighan, Rewon
  Child, Aditya Ramesh, Daniel Ziegler, Jeffrey Wu, Clemens Winter, Chris
  Hesse, Mark Chen, Eric Sigler, Mateusz Litwin, Scott Gray, Benjamin Chess,
  Jack Clark, Christopher Berner, Sam McCandlish, Alec Radford, Ilya Sutskever,
  and Dario Amodei. 2020{\natexlab{a}}.
\newblock \href
  {https://proceedings.neurips.cc/paper_files/paper/2020/file/1457c0d6bfcb4967418bfb8ac142f64a-Paper.pdf}
  {Language models are few-shot learners}.
\newblock In \emph{Advances in Neural Information Processing Systems},
  volume~33, pages 1877--1901. Curran Associates, Inc.

\bibitem[{Brown et~al.(2020{\natexlab{b}})Brown, Mann, Ryder, Subbiah, Kaplan,
  Dhariwal, Neelakantan, Shyam, Sastry, Askell et~al.}]{brown2020language}
Tom Brown, Benjamin Mann, Nick Ryder, Melanie Subbiah, Jared~D Kaplan, Prafulla
  Dhariwal, Arvind Neelakantan, Pranav Shyam, Girish Sastry, Amanda Askell,
  et~al. 2020{\natexlab{b}}.
\newblock Language models are few-shot learners.
\newblock \emph{Advances in neural information processing systems},
  33:1877--1901.

\bibitem[{Burtsev et~al.(2020)Burtsev, Kuratov, Peganov, and
  Sapunov}]{burtsev2020memory}
Mikhail~S Burtsev, Yuri Kuratov, Anton Peganov, and Grigory~V Sapunov. 2020.
\newblock Memory transformer.
\newblock \emph{arXiv preprint arXiv:2006.11527}.

\bibitem[{Choromanski et~al.(2021)Choromanski, Likhosherstov, Dohan, Song,
  Gane, Sarlos, Hawkins, Davis, Mohiuddin, Kaiser, Belanger, Colwell, and
  Weller}]{choromanski2021rethinking}
Krzysztof~Marcin Choromanski, Valerii Likhosherstov, David Dohan, Xingyou Song,
  Andreea Gane, Tamas Sarlos, Peter Hawkins, Jared~Quincy Davis, Afroz
  Mohiuddin, Lukasz Kaiser, David~Benjamin Belanger, Lucy~J Colwell, and Adrian
  Weller. 2021.
\newblock \href {https://openreview.net/forum?id=Ua6zuk0WRH} {Rethinking
  attention with performers}.
\newblock In \emph{International Conference on Learning Representations}.

\bibitem[{Dao(2023)}]{dao2023flashattention}
Tri Dao. 2023.
\newblock Flashattention-2: Faster attention with better parallelism and work
  partitioning.
\newblock \emph{arXiv preprint arXiv:2307.08691}.

\bibitem[{Dao et~al.(2022)Dao, Fu, Ermon, Rudra, and
  Re}]{dao2022flashattention}
Tri Dao, Daniel~Y Fu, Stefano Ermon, Atri Rudra, and Christopher Re. 2022.
\newblock \href {https://openreview.net/forum?id=H4DqfPSibmx} {Flashattention:
  Fast and memory-efficient exact attention with {IO}-awareness}.
\newblock In \emph{Advances in Neural Information Processing Systems}.

\bibitem[{Dasigi et~al.(2021)Dasigi, Lo, Beltagy, Cohan, Smith, and
  Gardner}]{dasigi-etal-2021-dataset}
Pradeep Dasigi, Kyle Lo, Iz~Beltagy, Arman Cohan, Noah~A. Smith, and Matt
  Gardner. 2021.
\newblock \href {https://doi.org/10.18653/v1/2021.naacl-main.365} {A dataset of
  information-seeking questions and answers anchored in research papers}.
\newblock In \emph{Proceedings of the 2021 Conference of the North American
  Chapter of the Association for Computational Linguistics: Human Language
  Technologies}, pages 4599--4610, Online. Association for Computational
  Linguistics.

\bibitem[{Devlin et~al.(2019)Devlin, Chang, Lee, and
  Toutanova}]{devlin-etal-2019-bert}
Jacob Devlin, Ming-Wei Chang, Kenton Lee, and Kristina Toutanova. 2019.
\newblock \href {https://doi.org/10.18653/v1/N19-1423} {{BERT}: Pre-training of
  deep bidirectional transformers for language understanding}.
\newblock In \emph{Proceedings of the 2019 Conference of the North {A}merican
  Chapter of the Association for Computational Linguistics: Human Language
  Technologies, Volume 1 (Long and Short Papers)}, pages 4171--4186,
  Minneapolis, Minnesota. Association for Computational Linguistics.

\bibitem[{Fabbri et~al.(2021)Fabbri, Kryściński, McCann, Xiong, Socher, and
  Radev}]{tacla00373}
Alexander~R. Fabbri, Wojciech Kryściński, Bryan McCann, Caiming Xiong,
  Richard Socher, and Dragomir Radev. 2021.
\newblock \href {https://doi.org/10.1162/tacl_a_00373} {{SummEval:
  Re-evaluating Summarization Evaluation}}.
\newblock \emph{Transactions of the Association for Computational Linguistics},
  9:391--409.

\bibitem[{Huang et~al.(2021)Huang, Cao, Parulian, Ji, and
  Wang}]{huang-etal-2021-efficient}
Luyang Huang, Shuyang Cao, Nikolaus Parulian, Heng Ji, and Lu~Wang. 2021.
\newblock \href {https://doi.org/10.18653/v1/2021.naacl-main.112} {Efficient
  attentions for long document summarization}.
\newblock In \emph{Proceedings of the 2021 Conference of the North American
  Chapter of the Association for Computational Linguistics: Human Language
  Technologies}, pages 1419--1436, Online. Association for Computational
  Linguistics.

\bibitem[{Hutchins et~al.(2022)Hutchins, Schlag, Wu, Dyer, and
  Neyshabur}]{hutchins2022blockrecurrent}
DeLesley Hutchins, Imanol Schlag, Yuhuai Wu, Ethan Dyer, and Behnam Neyshabur.
  2022.
\newblock \href {https://openreview.net/forum?id=uloenYmLCAo} {Block-recurrent
  transformers}.
\newblock In \emph{Advances in Neural Information Processing Systems}.

\bibitem[{Ivgi et~al.(2023)Ivgi, Shaham, and Berant}]{ivgi-etal-2023-efficient}
Maor Ivgi, Uri Shaham, and Jonathan Berant. 2023.
\newblock \href {https://doi.org/10.1162/tacl_a_00547} {Efficient long-text
  understanding with short-text models}.
\newblock \emph{Transactions of the Association for Computational Linguistics},
  11:284--299.

\bibitem[{Kitaev et~al.(2020)Kitaev, Kaiser, and Levskaya}]{Kitaev2020Reformer}
Nikita Kitaev, Lukasz Kaiser, and Anselm Levskaya. 2020.
\newblock \href {https://openreview.net/forum?id=rkgNKkHtvB} {Reformer: The
  efficient transformer}.
\newblock In \emph{International Conference on Learning Representations}.

\bibitem[{Ko{\v{c}}isk{\'y} et~al.(2018)Ko{\v{c}}isk{\'y}, Schwarz, Blunsom,
  Dyer, Hermann, Melis, and Grefenstette}]{kocisky-etal-2018-narrativeqa}
Tom{\'a}{\v{s}} Ko{\v{c}}isk{\'y}, Jonathan Schwarz, Phil Blunsom, Chris Dyer,
  Karl~Moritz Hermann, G{\'a}bor Melis, and Edward Grefenstette. 2018.
\newblock \href {https://doi.org/10.1162/tacl_a_00023} {The {N}arrative{QA}
  reading comprehension challenge}.
\newblock \emph{Transactions of the Association for Computational Linguistics},
  6:317--328.

\bibitem[{Kryscinski et~al.(2022)Kryscinski, Rajani, Agarwal, Xiong, and
  Radev}]{kryscinski-etal-2022-booksum}
Wojciech Kryscinski, Nazneen Rajani, Divyansh Agarwal, Caiming Xiong, and
  Dragomir Radev. 2022.
\newblock \href {https://doi.org/10.18653/v1/2022.findings-emnlp.488}
  {{BOOKSUM}: A collection of datasets for long-form narrative summarization}.
\newblock In \emph{Findings of the Association for Computational Linguistics:
  EMNLP 2022}, pages 6536--6558, Abu Dhabi, United Arab Emirates. Association
  for Computational Linguistics.

\bibitem[{Lewis et~al.(2020)Lewis, Liu, Goyal, Ghazvininejad, Mohamed, Levy,
  Stoyanov, and Zettlemoyer}]{lewis-etal-2020-bart}
Mike Lewis, Yinhan Liu, Naman Goyal, Marjan Ghazvininejad, Abdelrahman Mohamed,
  Omer Levy, Veselin Stoyanov, and Luke Zettlemoyer. 2020.
\newblock \href {https://doi.org/10.18653/v1/2020.acl-main.703} {{BART}:
  Denoising sequence-to-sequence pre-training for natural language generation,
  translation, and comprehension}.
\newblock In \emph{Proceedings of the 58th Annual Meeting of the Association
  for Computational Linguistics}, pages 7871--7880, Online. Association for
  Computational Linguistics.

\bibitem[{Lin(2004)}]{lin-2004-rouge}
Chin-Yew Lin. 2004.
\newblock \href {https://aclanthology.org/W04-1013} {{ROUGE}: A package for
  automatic evaluation of summaries}.
\newblock In \emph{Text Summarization Branches Out}, pages 74--81, Barcelona,
  Spain. Association for Computational Linguistics.

\bibitem[{Liu et~al.(2020)Liu, Ott, Goyal, Du, Joshi, Chen, Levy, Lewis,
  Zettlemoyer, and Stoyanov}]{liu2020roberta}
Yinhan Liu, Myle Ott, Naman Goyal, Jingfei Du, Mandar Joshi, Danqi Chen, Omer
  Levy, Mike Lewis, Luke Zettlemoyer, and Veselin Stoyanov. 2020.
\newblock \href {https://openreview.net/forum?id=SyxS0T4tvS} {Ro{\{}bert{\}}a:
  A robustly optimized {\{}bert{\}} pretraining approach}.

\bibitem[{Penedo et~al.(2023)Penedo, Malartic, Hesslow, Cojocaru, Cappelli,
  Alobeidli, Pannier, Almazrouei, and Launay}]{penedo2023refinedweb}
Guilherme Penedo, Quentin Malartic, Daniel Hesslow, Ruxandra Cojocaru,
  Alessandro Cappelli, Hamza Alobeidli, Baptiste Pannier, Ebtesam Almazrouei,
  and Julien Launay. 2023.
\newblock The refinedweb dataset for falcon llm: outperforming curated corpora
  with web data, and web data only.
\newblock \emph{arXiv preprint arXiv:2306.01116}.

\bibitem[{Peng et~al.(2023)Peng, Alcaide, Anthony, Albalak, Arcadinho, Cao,
  Cheng, Chung, Grella, GV et~al.}]{peng2023rwkv}
Bo~Peng, Eric Alcaide, Quentin Anthony, Alon Albalak, Samuel Arcadinho, Huanqi
  Cao, Xin Cheng, Michael Chung, Matteo Grella, Kranthi~Kiran GV, et~al. 2023.
\newblock Rwkv: Reinventing rnns for the transformer era.
\newblock \emph{arXiv preprint arXiv:2305.13048}.

\bibitem[{Poli et~al.(2023)Poli, Massaroli, Nguyen, Fu, Dao, Baccus, Bengio,
  Ermon, and R{\'e}}]{poli2023hyena}
Michael Poli, Stefano Massaroli, Eric Nguyen, Daniel~Y Fu, Tri Dao, Stephen
  Baccus, Yoshua Bengio, Stefano Ermon, and Christopher R{\'e}. 2023.
\newblock Hyena hierarchy: Towards larger convolutional language models.
\newblock \emph{arXiv preprint arXiv:2302.10866}.

\bibitem[{Press et~al.(2022)Press, Smith, and Lewis}]{press2022train}
Ofir Press, Noah Smith, and Mike Lewis. 2022.
\newblock \href {https://openreview.net/forum?id=R8sQPpGCv0} {Train short, test
  long: Attention with linear biases enables input length extrapolation}.
\newblock In \emph{International Conference on Learning Representations}.

\bibitem[{Rubin and Berant(2023)}]{rubin2023long}
Ohad Rubin and Jonathan Berant. 2023.
\newblock Long-range language modeling with self-retrieval.
\newblock \emph{arXiv preprint arXiv:2306.13421}.

\bibitem[{Shaham et~al.(2023)Shaham, Ivgi, Efrat, Berant, and
  Levy}]{shaham2023zeroscrolls}
Uri Shaham, Maor Ivgi, Avia Efrat, Jonathan Berant, and Omer Levy. 2023.
\newblock Zeroscrolls: A zero-shot benchmark for long text understanding.
\newblock \emph{arXiv preprint arXiv:2305.14196}.

\bibitem[{Su et~al.(2021)Su, Lu, Pan, Wen, and Liu}]{su2021roformer}
Jianlin Su, Yu~Lu, Shengfeng Pan, Bo~Wen, and Yunfeng Liu. 2021.
\newblock \href {http://arxiv.org/abs/2104.09864} {Roformer: Enhanced
  transformer with rotary position embedding}.

\bibitem[{Sun et~al.(2023)Sun, Dong, Patra, Ma, Huang, Benhaim, Chaudhary,
  Song, and Wei}]{sun-etal-2023-length}
Yutao Sun, Li~Dong, Barun Patra, Shuming Ma, Shaohan Huang, Alon Benhaim,
  Vishrav Chaudhary, Xia Song, and Furu Wei. 2023.
\newblock \href {https://doi.org/10.18653/v1/2023.acl-long.816} {A
  length-extrapolatable transformer}.
\newblock In \emph{Proceedings of the 61st Annual Meeting of the Association
  for Computational Linguistics (Volume 1: Long Papers)}, pages 14590--14604,
  Toronto, Canada. Association for Computational Linguistics.

\bibitem[{Tay et~al.(2022)Tay, Dehghani, Bahri, and Metzler}]{10.1145/3530811}
Yi~Tay, Mostafa Dehghani, Dara Bahri, and Donald Metzler. 2022.
\newblock \href {https://doi.org/10.1145/3530811} {Efficient transformers: A
  survey}.
\newblock \emph{ACM Comput. Surv.}, 55(6).

\bibitem[{Team(2023)}]{MosaicML2023Introducing}
MosaicML~NLP Team. 2023.
\newblock \href {www.mosaicml.com/blog/mpt-7b} {Introducing mpt-7b: A new
  standard for open-source, commercially usable llms}.
\newblock Accessed: 2023-10-05.

\bibitem[{Touvron et~al.(2023)Touvron, Martin, Stone, Albert, Almahairi,
  Babaei, Bashlykov, Batra, Bhargava, Bhosale et~al.}]{touvron2023llama}
Hugo Touvron, Louis Martin, Kevin Stone, Peter Albert, Amjad Almahairi, Yasmine
  Babaei, Nikolay Bashlykov, Soumya Batra, Prajjwal Bhargava, Shruti Bhosale,
  et~al. 2023.
\newblock Llama 2: Open foundation and fine-tuned chat models.
\newblock \emph{arXiv preprint arXiv:2307.09288}.

\bibitem[{Wang et~al.(2020)Wang, Li, Khabsa, Fang, and Ma}]{wang2020linformer}
Sinong Wang, Belinda~Z Li, Madian Khabsa, Han Fang, and Hao Ma. 2020.
\newblock Linformer: Self-attention with linear complexity.
\newblock \emph{arXiv preprint arXiv:2006.04768}.

\bibitem[{Yang et~al.(2019)Yang, Dai, Yang, Carbonell, Salakhutdinov, and
  Le}]{yang2019xlnet}
Zhilin Yang, Zihang Dai, Yiming Yang, Jaime Carbonell, Russ~R Salakhutdinov,
  and Quoc~V Le. 2019.
\newblock Xlnet: Generalized autoregressive pretraining for language
  understanding.
\newblock \emph{Advances in neural information processing systems}, 32.

\bibitem[{Zaheer et~al.(2020)Zaheer, Guruganesh, Dubey, Ainslie, Alberti,
  Ontanon, Pham, Ravula, Wang, Yang et~al.}]{zaheer2020big}
Manzil Zaheer, Guru Guruganesh, Kumar~Avinava Dubey, Joshua Ainslie, Chris
  Alberti, Santiago Ontanon, Philip Pham, Anirudh Ravula, Qifan Wang, Li~Yang,
  et~al. 2020.
\newblock Big bird: Transformers for longer sequences.
\newblock \emph{Advances in neural information processing systems},
  33:17283--17297.

\bibitem[{Zhang* et~al.(2020)Zhang*, Kishore*, Wu*, Weinberger, and
  Artzi}]{bert-score}
Tianyi Zhang*, Varsha Kishore*, Felix Wu*, Kilian~Q. Weinberger, and Yoav
  Artzi. 2020.
\newblock \href {https://openreview.net/forum?id=SkeHuCVFDr} {Bertscore:
  Evaluating text generation with bert}.
\newblock In \emph{International Conference on Learning Representations}.

\end{thebibliography}

\appendix

\section{Implementation Comparison}
\label{app:unlim}
Although similar in using separate indices for each layer, we present additional modifications to the original methodology\footnote{\url{https://github.com/abertsch72/unlimiformer}}.
Specifically, 1) we introduce an update procedure for indices to avoid staleness, 2) we use a slightly more efficient chunk encoding approach, and 3) we introduce information fusion into the architecture and reformulate the attention calculations to be more comprehensive (see Section \ref{sec:meth:ca}).
Moreover, to the best of our knowledge, no evaluation of their proposed methodology has been presented for decoder-only models, a shortcoming that our work aims to address using a new evaluation setting and new datasets.

\section{Hyperparameters}
\label{app:hyper}
\paragraph{Cross-Attention Layers ($\mathcal{L}$)}
To find the best $\mathcal{L}$, first, we find the highest-performing single-layer pattern.
Then, we expand that pattern by adding one more cross-attention layer before or after that layer, with varying distances, until no performance improvement can be seen over the validation set.
We continue this process up to three layers.
For example, if $\mathcal{L} = \{ 16\}$ is the highest-performing single-layer pattern, at the second step, we will try $\mathcal{L} \in \{\{ 16, 17\}, \{ 15, 16\}, \{ 16, 18\}, \hdots\}$, and then at the third step, assuming $\mathcal{L} = \{ 16, 18\}$ was the best-performing pattern, we will try $\mathcal{L} \in \{\{ 14, 16, 18\}, \{ 16, 18, 20\}, \hdots\}$.
In most of our experiments, the best performance was achieved by $\mathcal{L} = \{ 20, 22, 24\}$.

\paragraph{Retention Sizes ($\alpha_q, \beta_q, \beta_{kv}, \alpha_{kv}$)}
After tuning $\mathcal{L}$, we tune these hyperparameters by first constraining them with $\alpha_q + \beta_q = \beta_{kv} + \alpha_{kv} = S$, where $S$ is the context length, and then doing a sweep on $\alpha_q, \alpha_{kv} \in \{0.05, 0.1, 0.15, \hdots, 0.95, 1\} \times S$.
In most of our experiments, the best performance was achieved by $\alpha_q, \alpha_{kv} = 0.4 \times S$.

\end{document}